\documentclass[preprint,3p,10pt,times]{elsarticle}

\usepackage[T1]{fontenc}
\usepackage{mathptmx}
\usepackage{amsmath, amssymb, amsthm}
\usepackage{algorithm}
\usepackage{algpseudocode}
\usepackage{array}
\usepackage{booktabs}
\usepackage{graphicx}
\usepackage[font=footnotesize,labelfont=bf]{caption}
\usepackage{hyperref}
\usepackage{cleveref}

\newtheorem{theorem}{Theorem}
\newtheorem*{theorem*}{Theorem}

\begin{document}

\begin{frontmatter}

\journal{Pattern Recognition}

\title{Partition-Invariant Federated Class-Incremental Learning via Closed-Form Aggregation of Sufficient Statistics}

\author[udc]{Alejandro Dopico-Castro\corref{cor1}}
\ead{alejandro.dopico2@udc.es}
\author[udc]{Oscar Fontenla-Romero}
\ead{oscar.fontenla@udc.es}
\author[udc]{Bertha Guijarro-Berdiñas}
\ead{berta.guijarro@udc.es}
\author[udc]{Amparo Alonso-Betanzos}
\ead{amparo.alonso.betanzos@udc.es}

\cortext[cor1]{\textit{Corresponding author.}}

\affiliation[udc]{
    organization={Universidade da Coruña, CITIC},
    addressline={Facultade de Informática, Campus de Elviña s/n}, 
    city={A Coruña},
    postcode={15071}, 
    state={Galicia},
    country={Spain}
}

\begin{abstract}
Federated Class-Incremental Learning (FedCIL) requires a model to recognize an expanding set of visual categories while its training data remain distributed across clients that cannot share them. Most existing methods treat this as an iterative optimization problem, repeatedly fine-tuning the network and averaging updates whenever new classes appear, which is costly and makes the global model sensitive to the client data distribution. Strong pretrained visual representations raise a more basic question: if such a representation is kept fixed, does learning the classifier still require iterative federated optimization? We show that it does not. With fixed features, classifier learning becomes a closed-form statistical estimation problem: clients share compact summaries of their data instead of gradients, and the server combines them into the same classifier that centralized training on that data would produce. A separate difficulty is specific to the incremental setting: classifiers of newly introduced classes receive no evidence from previously learned ones. We address it with a replay of stored feature embeddings coordinated across clients, without storing or transmitting raw images. We instantiate both ideas in FedIAL, a \textbf{Fed}erated and \textbf{I}ncremental \textbf{A}nalytic class \textbf{L}earning algorithm. On CIFAR-100 and TinyImageNet, FedIAL improves final accuracy by 5.6 and 13.8 points over the strongest baseline, with a single synchronization cycle per task and over 90\% less energy than generative-replay methods. On the domain-shifted ImageNet-R, a single, bounded adaptation of the representation before incremental learning recovers nearly all the accuracy of continuous adaptation (72.00 vs.\ 72.20) at a quarter of the energy.
\end{abstract}


\end{frontmatter}

\section{Introduction}
\label{sec:Introduction}

Federated Learning (FL) and Class-Incremental Learning (CIL) address complementary challenges: FL handles decentralized private data across multiple clients, while CIL manages the temporal arrival of new data categories. Many real-world systems, such as autonomous fleets or mobile healthcare, face both simultaneously. This combination, Federated Class-Incremental Learning (FedCIL), is particularly difficult because it mixes the spatial heterogeneity of Non-Independent and Identically Distributed (Non-IID) client data with the temporal arrival of new classes, creating an unstable learning problem prone to catastrophic forgetting and representation drift \cite{van2022three}.

In CIL, learning is typically organized into a sequence of tasks, each introducing a disjoint set of new classes that must be integrated without forgetting previous ones. Existing FedCIL methods counter forgetting in different ways, through regularization and distillation, generative replay of past classes, or adaptation of pretrained backbones, but they share an assumption inherited from federated learning \cite{mcmahan_communication-efficient_2017}, namely that learning proceeds by iterative gradient-based optimization, with repeated local training and averaging of updates every time new classes arrive. This is costly, and it ties the quality of the global model to how data happen to be distributed across clients, since averaged updates do not in general recover what a comparable centralized training would produce.

Pretrained visual representations change this picture. When a backbone already provides transferable features \cite{chen2020simple,he2020momentum}, much of the learning problem shifts to the classifier placed on top of it, which raises the question this paper is organized around: \emph{once a sufficiently transferable representation is available and kept fixed, does classifier learning in FedCIL still require iterative federated optimization?} Analytic learning suggests it may not. In centralized CIL, closed-form classifiers over frozen features have matched gradient-based training at a fraction of the cost \cite{zhuang2022acil}, and STSA (Spatial-Temporal Statistics Aggregation) \cite{guan2025stsa} has begun to bring this idea to the federated setting by aggregating feature statistics across clients and tasks into a closed-form classifier, although it still relies on gradient-based training in its first task. In this paper, we show that, for fixed features, the classifier can be obtained by solving a convex optimization problem fully determined by compact statistics of each client's data, namely second-order summaries of its features and targets. In our proposal, clients send these statistics instead of gradients, and the server combines them to obtain the same classifier that centralized training on that same data would produce, regardless of how the data are partitioned in the federated setting. Federation changes \emph{where} the statistics are computed, not \emph{how} the classifier is learned.

This does not, by itself, solve the incremental part of the problem. When a classifier is learned only from the data currently available, as in our formulation, the output units of newly arrived classes never observe samples of earlier classes, so they lack the negative evidence needed to separate new categories from old ones. This difficulty is independent of how the classifier is computed, since it stems from missing data rather than from the optimizer. 

We consider these two problems separately throughout the paper and propose FedIAL (Federated and Incremental Analytic class Learning), a federated learning algorithm which addresses each problem with its own mechanism: a frozen pretrained backbone shared by all clients, a closed-form classifier computed from aggregated statistics from clients, and a coordinated replay of stored feature embeddings, balanced across the federation, that supplies the historical evidence new classes lack without storing raw images. Because the approach presumes a transferable representation, we also examine what happens when this premise weakens under domain shift, and show that a single, bounded adaptation of the representation before incremental learning begins is enough for the same analytic formulation to apply again.

The main contributions of this work are the following:

\begin{itemize}
    \item A reformulation of classifier learning in FedCIL under a fixed pretrained representation as a closed-form, distributed sufficient-statistics problem, which removes the need for iterative classifier optimization in this regime.
    \item An aggregation scheme in which clients exchange sufficient statistics instead of gradients. We prove that the global classifier equals the centralized Regularized Recursive Least Squares (RRLS) solution over the same client-side data, under exact arithmetic and without lossy rank truncation, independently of data partitioning, aggregation order, and communication topology.
    \item The identification of missing historical evidence as a separate problem for newly introduced classes, and a globally coordinated latent replay protocol that addresses it without sharing raw data.
    \item An empirical study on CIFAR-100 and TinyImageNet across different heterogeneity levels, client pool sizes, and task granularity, characterizing the accuracy-efficiency trade-off against gradient-based and analytic baselines, and showing that FedIAL attains the highest final accuracy in all evaluated settings, remains stable under severe heterogeneity, and consumes over 90\% less energy than generative-replay baselines.
    \item A controlled relaxation of the fixed-representation assumption on ImageNet-R, showing that one bounded adaptation step before the incremental phase restores the conditions for analytic classifier learning.
\end{itemize}

The remainder of this work is organized as follows: \Cref{sec:related} reviews related work in Federated Learning, Continual Learning, and their intersection. \Cref{sec:preliminary} introduces the necessary preliminaries and problem formulation. \Cref{sec:method} presents the proposed method, including the analytic local solver and the aggregation protocol. \Cref{sec:experiments} reports experimental results and ablation studies. \Cref{sec:conclusion} concludes the paper and discusses limitations and future directions.

\section{Related Work}
\label{sec:related}

The challenge of Federated Class-Incremental Learning (FedCIL) stems from the compounding effects of spatial data heterogeneity and temporal expansion of the label space, often referred to as the ``compound forgetting effect'' \cite{shenaj2023asynchronous,ma2022continual}.

Early FL works (FedAvg \cite{mcmahan_communication-efficient_2017}, FedProx \cite{li2020federated}) assumed a static label space. Standard CIL methods, replay \cite{Rebuffi2017, lopez2017gradient,buzzega2020dark}, regularization \cite{li2017learning, zenke2017continual}, and architecture isolation \cite{mallya2018packnet}, address the stability-plasticity dilemma \cite{kirkpatrick2017overcoming}, the trade-off between learning new information and preserving old knowledge, but do not port trivially to federated environments. Standard functional distillation, where the current local model is regularized to mimic the previous global model's outputs, collapses under Non-IID data \cite{wang2020tackling, hsu2019measuring}, as local gradients diverge from the global objective, causing the ``student'' model to drift toward a biased local optimum rather than the global consensus. More recent approaches seek to alleviate this feature drift through refined distillation strategies. FedCAD~\cite{ZHANG2026109298} addresses cross-domain heterogeneous FL via cross-modal semantic alignment and distillation, while FedSSD~\cite{he2022learning} employs a selective self-distillation strategy to stabilize local representations without external replay. Although these methods mitigate feature drift and representation collapse, they remain designed for static or cross-domain settings, leaving the temporal expansion of class-incremental scenarios largely unaddressed.

Recent work in FedCIL focuses on reconciling local updates with global consistency to ensure decentralized learning does not fragment the representation space. GLFC \cite{dong_federated_2022} introduces a class-gradient compensation mechanism to balance local and global proxies. To avoid the privacy risks of raw image replay, TARGET \cite{zhang_target_2023} uses a global model to guide generative replay, decoupling knowledge transfer from local data. Similarly, LANDER \cite{tran_text-enhanced_2024} leverages pretrained language embeddings as semantic anchors to structure the latent space and mitigate non-anchored representations in data-free settings. These methods, however, still rely on iterative generative optimization or expensive backpropagation, incurring significant computational and communication overhead. Our method avoids this generative complexity by employing memory-efficient latent embeddings and a closed-form solver.

To address these inefficiencies, a growing branch of research seeks to replace iterative fine-tuning with Analytic Learning solutions \cite{zhuang2023gkeal,zhuang2024ds}. In centralized CIL, it has been shown that frozen, high-quality backbones coupled with analytical heads can match gradient-based performance with substantially lower training time \cite{zhuang2022acil,dopico2025efficient, Gao_2025_CVPR, zhou2025revisiting}.

Extending this analytical paradigm to Federated CIL, frameworks such as STSA \cite{guan2025stsa} have begun aggregating spatiotemporal statistics to unify learning. Our work follows the same analytical philosophy while enforcing representation invariance through a frozen backbone from the first task onward, enabling memory-efficient latent replay and closed-form classifier updates. Unlike STSA, which relies on gradient-based optimization during the initial task before switching to analytic updates, FedIAL performs no iterative classifier optimization at any task under the standard fixed-representation regime, relying instead on closed-form aggregation of sufficient statistics throughout. This property allows the server to perform aggregation via Regularized Recursive Least Squares (RRLS) \cite{zhuang2022acil,fontenla2023fedheonn} that is, under certain conditions, an exact fusion of local sufficient statistics rather than a stochastic approximation. FedIAL further introduces a globally coordinated replay mechanism that is absent from STSA. Beyond analytic aggregation, other recent work adapts ViT backbones directly for FedCIL. PILoRA~\cite{guo2024pilora} combines prototype learning with incremental LoRA modules encoded per task to jointly address catastrophic forgetting and client heterogeneity, while FCILPT~\cite{luo2025federated} is a rehearsal-free approach that encodes task-relevant and task-irrelevant knowledge into aligned prompt pools across clients. Relative to this line of work, our contribution is a precise formulation and centralized-equivalence guarantee for classifier learning under the fixed-representation regime in the federated class-incremental setting, together with an explicit separation between this optimization question and the distinct problem of missing historical evidence for newly introduced classes, which coordinated latent replay is designed to address. Finally, we also provide a solution for a controlled, adapter-based relaxation of the fixed-representation assumption itself, for settings where it does not hold.

\section{Problem Formalization}
\label{sec:preliminary}

In the Federated Class-Incremental Learning (FedCIL) paradigm, we consider a global model $f_\Theta$ parameterized by $\Theta$, which is collaboratively trained across a set of $N$ distributed clients over a temporal sequence of $T$ tasks. Each task $t \in \{1, \ldots, T\}$ introduces a disjoint set of classes $\mathcal{C}_t$, such that $\mathcal{C}_t \cap \mathcal{C}_{t'} = \emptyset,\ \forall\, t \neq t'$. At any given task $t$, each client $i$ gains access to a local dataset $\mathcal{D}_{i,t} = \{(x_j, y_j) \mid y_j \in \mathcal{C}_{i,t}\}$, where the local label space $\mathcal{C}_{i,t}$ is a subset of the global task space $\mathcal{C}_t$. Accordingly, the dataset associated with task $t$ is defined as $\mathcal{D}_t = \bigcup_{i=1}^{N}\mathcal{D}_{i,t}$. The objective is to obtain an optimal global parameter state $\Theta^*$ that minimizes the empirical risk over the cumulative distribution of all tasks encountered so far, $\mathcal{D}_{1:T} = \bigcup_{t=1}^{T} \mathcal{D}_t$, without requiring the retention of raw data from previous tasks. This is formally expressed as:

\begin{equation}
    \Theta^* = \arg\min_{\Theta} \sum_{t=1}^{T} \mathbb{E}_{(x,y) \sim \mathcal{D}_t} [\ell(f_\Theta(x), y)]
    \label{eq:global_objective}
\end{equation}

\noindent where $\ell(\cdot)$ denotes a generic loss function, typically the cross-entropy loss for classification tasks.

The formulation in Eq.~(\ref{eq:global_objective}) treats $\Theta$ as a single, undifferentiated parameter vector, but in practice $f_\Theta$ decomposes into a feature extractor and a classification head. We make this decomposition explicit by writing $f_\Theta(x) = W^\top F_\theta(x)$ with $\Theta = (\theta, W)$, where $F_\theta$ maps inputs to a $d$-dimensional embedding and $W$ is a linear classifier applied to that embedding, i.e., a single-layer network with one output unit per class. When $F_\theta$ is held fixed throughout the task sequence, the outer minimization over $\Theta$ in Eq.~(\ref{eq:global_objective}) reduces to a minimization over $W$ alone. This is the regime analyzed throughout the remainder of this paper. We will clarify both the benefits of this reduction in the federated setting and the limitations it leaves unaddressed. In this regime, we learn $W$ through a regularized least-squares surrogate of $\ell$ (\Cref{subsec:local}), yielding a convex problem with a closed-form solution. All equivalence statements below refer to this surrogate objective.

\begin{figure}[t]
    \centering
    \includegraphics[width=\linewidth]{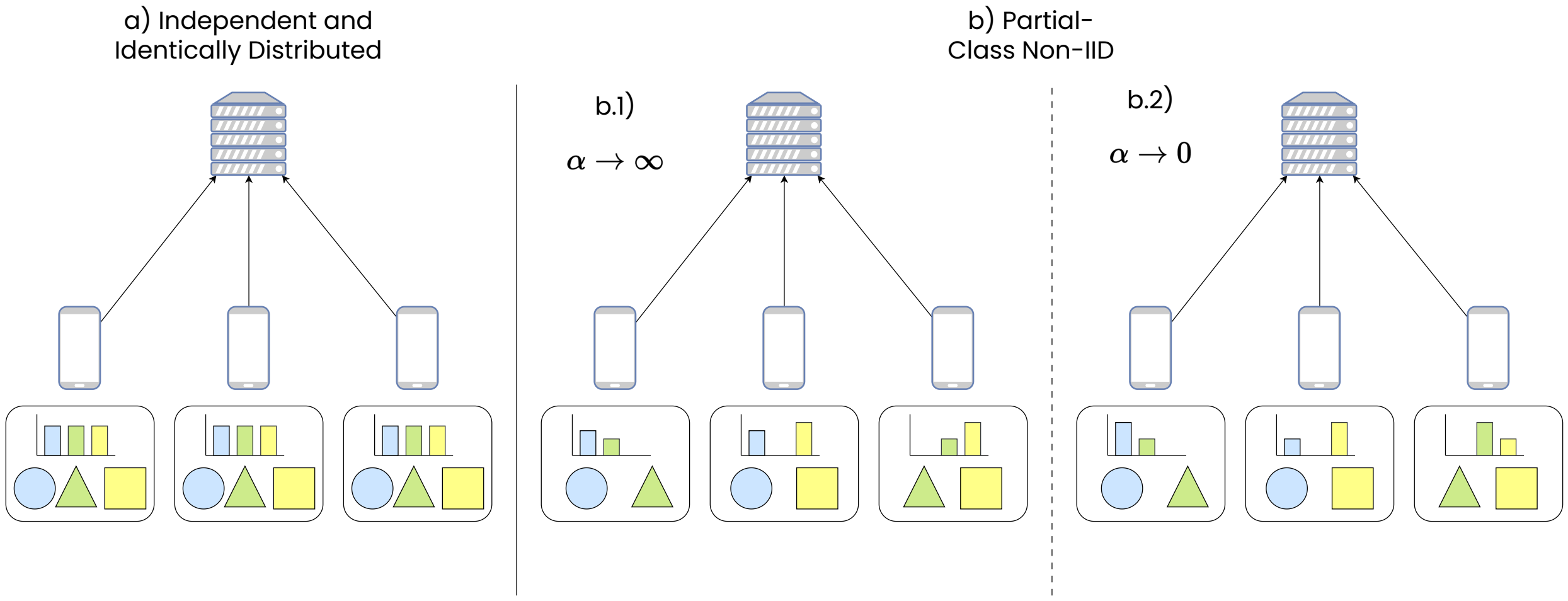}
    \caption{Statistical Heterogeneity in FL. (a) IID, clients sample from the full distribution $\mathcal{D}$. (b) Partial-Class (Non-IID), local label spaces are restricted ($\mathcal{C}_i \subset \mathcal{C}$), with each client assigned a fraction $\rho$ of classes and each shared class split across the clients that hold it according to $\mathrm{Dir}(\alpha)$ (small $\alpha$ concentrates it on few clients).}
    \label{fig:data_distribution}
\end{figure}

Achieving this objective is challenging due to the inherent statistical heterogeneity across the federation. While many federated learning methods are analyzed under the assumption of Independent and Identically Distributed (IID) data, where $\mathcal{D}_i = \mathcal{D}$ for all clients $i$, real-world FedCIL scenarios exhibit Partial-Class (Non-IID) distributions. In this setting, each client observes only a fraction $\rho$ of the classes available at each task. To model this heterogeneity, sample proportions are generated according to a Dirichlet distribution $\mathrm{Dir}(\alpha)$ \cite{hsu2019measuring, li2022federated}, which controls how evenly a shared class's samples are split across the clients that hold it; small $\alpha$ concentrates a class on a few clients, while $\alpha \to \infty$ recovers an even split. We illustrate this setting in \Cref{fig:data_distribution}.

This heterogeneity causes \emph{client drift} \cite{li2019convergence,karimireddy_scaffold_2020,acar2021federated}, local updates pulling the model toward disparate optima, and \emph{representation shift}, where the feature extractor adapts to new classes at the expense of prior semantic structure. Together, they amplify catastrophic forgetting and introduce class-imbalance bias toward recent classes, requiring a framework that balances stability with plasticity \cite{Rebuffi2017, kirkpatrick2017overcoming}.

Furthermore, the model must contend with severe \emph{class imbalance} and \emph{heterogeneous exposure}. In the former, the global model develops a prediction bias toward the most recently observed classes, leading to a collapse in performance on the ``long tail'' of historical data. In the latter, the disparate knowledge across clients, driven by the specific local partitions, results in a fragmented global view. When these locally biased models are aggregated, the resulting global model often suffers from significant variance, necessitating a framework that can maintain structural stability while ensuring plasticity for incoming tasks \cite{kirkpatrick2017overcoming, dong_federated_2022}.

\section{Proposed Method}
\label{sec:method}

\begin{figure}[t]
    \centering
    \includegraphics[width=\linewidth]{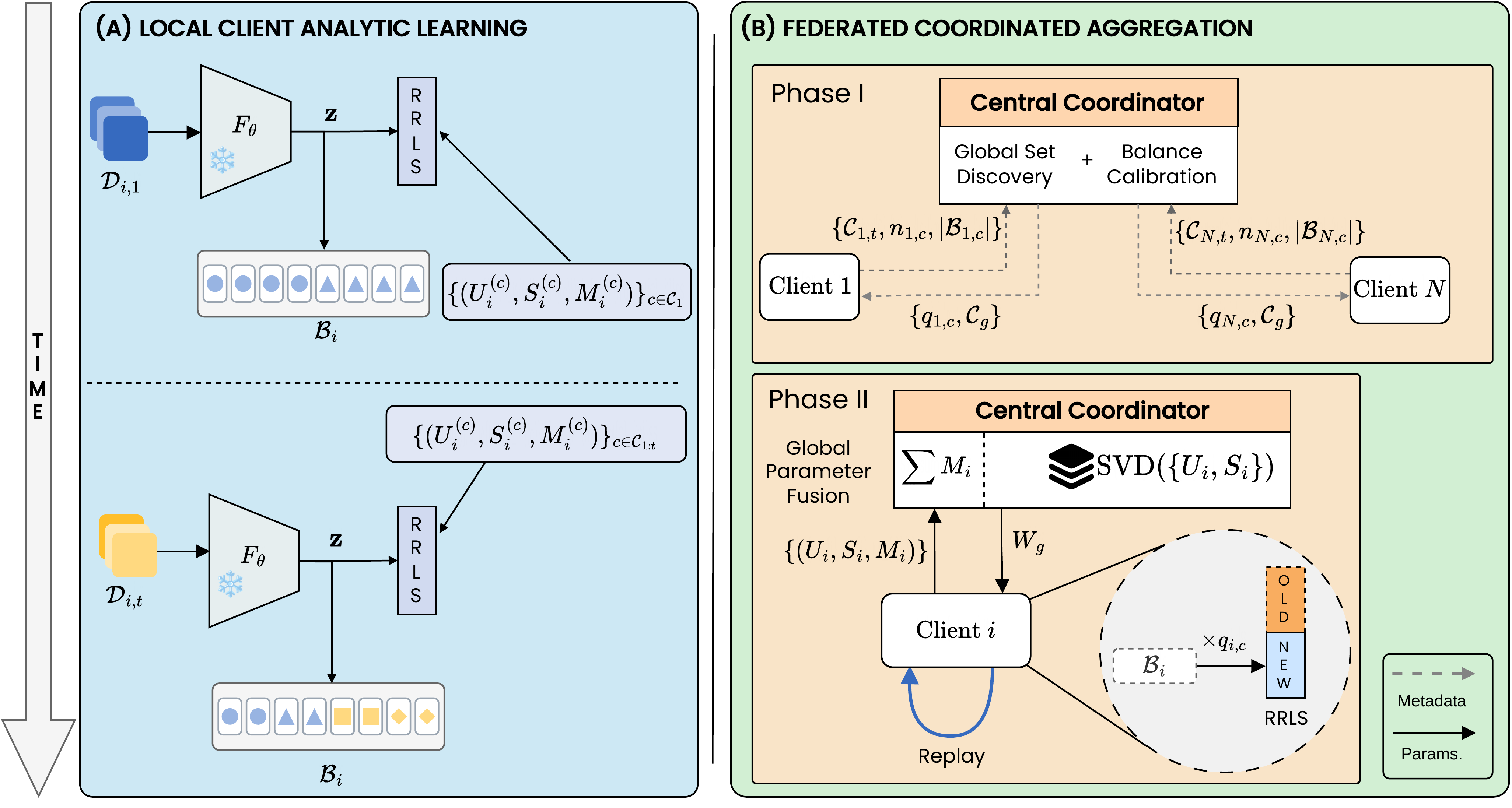}
    \caption{Overview of FedIAL. \textbf{(A) Local client analytic learning.} Each client feeds task data through the shared frozen backbone $F_\theta$, producing $d$-dimensional embeddings $\mathbf{z}$. These embeddings are used to update the analytic RRLS classifier, yielding per-class sufficient statistics $\{(U_i^{(c)}, S_i^{(c)}, M_i^{(c)})\}$, where $M_i^{(c)}$ is the $c$-th column of the moment matrix. A compact subset of embeddings is stored in the latent replay buffer $\mathcal{B}_i$ for use in future tasks. \textbf{(B) Federated coordinated aggregation.} \textit{Phase~I (metadata):} each client reports its local class set $\mathcal{C}_{i, t}$, sample counts $n_{i,c}$, and buffer sizes $|\mathcal{B}_{i, c}|$ to the coordinator, which computes global class statistics and broadcasts per-client oversampling quotas $q_{i,c}$ alongside the global class set $\mathcal{C}_g$. \textit{Phase~II (model parameters):} clients oversample $\mathcal{B}_i$ according to $q_{i,c}$, recompute statistics $(U_i, S_i, M_i)$ on the augmented data, and transmit them to the coordinator. The coordinator fuses them via $\sum M_i$ and $\mathrm{SVD}(\{U_i, S_i\})$ to produce global classifier $W_g$, which is broadcast back to all clients.}
    \label{fig:architecture}
\end{figure}

The central observation of this work is that, under a fixed, transferable representation, federated classifier learning reduces to a distributed sufficient-statistics problem, and FedIAL is the protocol that operationalizes it. On top of this, the class-incremental setting introduces one further challenge, the absence of historical evidence for newly introduced classes. The protocol rests on three pillars: (1) a stable, shared feature space that prevents representation drift across clients; (2) an analytic local solver that avoids iterative classifier optimization through closed-form updates; and (3) a coordinated global aggregation protocol that reconstructs the centralized classifier without exchanging raw data. These components resolve the optimization question posed in the Introduction. The second challenge, namely that the output units of new classes never observe samples of earlier classes, is addressed by a coordinated replay of stored embeddings embedded in the aggregation protocol. \Cref{fig:architecture} summarizes the protocol: part (A) corresponds to the local learning described in \Cref{subsec:local} and part (B) to the coordinated aggregation described in \Cref{subsec:aggregation}. 

\subsection{Local Client Analytic Learning}
\label{subsec:local}

\Cref{fig:architecture} (part A) illustrates the local pipeline executed by each client during every incremental task. It consists of a frozen feature extractor $F_\theta$ that maps inputs into a shared embedding space $\mathbb{R}^d$, and an analytic RRLS classifier that updates incrementally without backpropagation.

\subsubsection{Frozen backbone and embedding extraction}

Representation drift, caused by per-client backbone updates diverging across the federation, renders prior global weights incompatible with new features. To neutralize this drift, we fix a backbone $F_\theta$ pretrained on a large-scale proxy dataset \cite{chen2020big} and use it to project all inputs into a shared $d$-dimensional embedding space. For any client $i$ and input $x$, the embedding $\mathbf{z}$ is obtained as:
\begin{equation}
    \mathbf{z} = F_\theta(x), \quad \mathbf{z} \in \mathbb{R}^d
\end{equation}

By keeping $\theta$ frozen and identical across all clients, every client operates within the same semantic vector space throughout the entire incremental sequence of tasks. This has two important consequences. First, representational drift is structurally avoided, since the relative positions of class embeddings remain fixed, preserving earlier decision boundaries when new classes arrive. Second, feature extraction reduces to a single forward pass per sample with no gradient computation, substantially reducing local compute cost. A compact subset of these embeddings is stored in the client's latent replay buffer $\mathcal{B}_i$ for use in subsequent tasks.

\subsubsection{Analytic RRLS Classifier}

With the embedding space fixed, each client's optimization problem reduces to fitting a linear head over frozen features; local optimization thus becomes a convex ridge-regression task with a unique global optimum. Traditional federated incremental learning frameworks rely on iterative Stochastic Gradient Descent (SGD), which introduces optimization variability, client drift under Non-IID data, and hyperparameter sensitivity, aggravating catastrophic forgetting.

To overcome these limitations, FedIAL solves the local classifier optimization analytically using a \emph{Regularized Recursive Least Squares} (RRLS) solver \cite{zhuang2022acil,fontenla2023fedheonn}. Rather than iteratively approximating the decision boundary, RRLS computes the closed-form solution of the convex objective in a single pass without backpropagation. Beyond avoiding iterative classifier optimization, the RRLS formulation exposes a compact set of sufficient statistics $(U_i, S_i, M_i)$ that completely summarize the local data for this problem. This property is what makes iterative federated classifier optimization unnecessary, since the statistics can be aggregated globally, without communicating raw samples or intermediate gradients, and the centralized solution reconstructed from them.

When a new task $t$ arrives, the classification head is expanded to accommodate the new class set $\mathcal{C}_t$. Output neurons associated with previously learned classes $\mathcal{C}_{1:t-1}$ keep their accumulated statistics and are updated recursively with the current-task samples as negative evidence (target 0), whereas new neurons have no accumulated statistics and are estimated from current-task data only. Local RRLS learning proceeds in two stages: (i) mapping the targets to the pre-activation domain to construct a weighted ridge-regression problem alongside subspace decomposition, and (ii) extracting sufficient statistics required for subsequent global aggregation. All computations are performed independently and in parallel across output classes.

\paragraph{Pre-activation targets and subspace decomposition}

To obtain a closed-form least-squares solution despite the non-linear sigmoid activation $\sigma(\cdot)$, the classification problem is reformulated in the pre-activation domain. Given binary targets $Y_i \in \{0,1\}^{m_i \times |\mathcal{C}_{1:t}|}$, we first map them to corresponding pre-activation target values through the inverse-sigmoid (logit) function:

\begin{equation}
    \bar{D}_i = \sigma^{-1}(Y_i) \in \mathbb{R}^{m_i \times |\mathcal{C}_{1:t}|},
\end{equation}

\noindent where $\sigma^{-1}(\cdot)$ is applied element-wise after smoothing the binary targets to $\{\epsilon,1-\epsilon\}$, with $\epsilon \ll 1$, so that the logit remains finite.

Let $Z_i \in \mathbb{R}^{m_i \times d}$ denote the local embedding matrix of client $i$, obtained by passing its $m_i$ local samples through the frozen backbone $F_\theta$, i.e.\ row $j$ of $Z_i$ is $\mathbf{z}^\top_{i,j}$ with $\mathbf{z}_{i,j} = F_\theta (x_{i,j})$. This matrix is augmented with a bias column to form $X_i = [\mathbf{1}_{m_i}, Z_i] \in \mathbb{R}^{m_i \times (d+1)}$. Each sample is then weighted, for an output class $c \in \mathcal{C}_{1:t}$, by the factor $\omega(u) = u(1-u)$ evaluated at its target pre-activation, which produces the diagonal sample-weighting matrix:

\begin{equation}
    \mathbf{G}_i^{(c)} = \mathrm{diag} \bigl(\omega(\bar d_{i,1,c}), \ldots, \omega(\bar d_{i,m_i,c})\bigr) \in \mathbb{R}^{m_i \times m_i}
    \label{eq:G}
\end{equation}

\noindent where $\bar d_{i,j,c}$ is the entry of $\bar{D}_i$ for sample $j$ and class $c$. Since the smoothed targets take only two values, $\omega$ assigns one weight to the positive samples of class $c$ and another to its negatives. The weighting therefore depends on class membership, not on the position of a sample relative to the decision boundary. Because these weights depend on the targets of each output class, the subspace statistics are computed separately for each class $c$. To lighten notation, we omit the superscript $(c)$ from here on, so that all subsequent derivations apply class by class. The weighted design matrix $X_i^\top\mathbf{G}_i \in \mathbb{R}^{(d+1)\times m_i}$ is then factorized using Singular Value Decomposition (SVD):

\begin{equation}
    [U_i, S_i, \sim] = \mathrm{SVD}(X_i^\top\mathbf{G}_i),
\end{equation}

\noindent where $U_i \in \mathbb{R}^{(d+1) \times r_i}$ spans the principal subspace of the weighted features. This decomposition avoids explicit inversion of high-dimensional matrices and yields a numerically stable, regularized least-squares solution.

\paragraph{Sufficient statistics}

Alongside the SVD, client $i$ computes a moment matrix $M_i \in \mathbb{R}^{(d+1) \times |\mathcal{C}_{1:t}|}$ capturing cross-correlations between weighted features and target pre-activations, whose $c$-th column $M_i^{(c)}$ holds the moment of class $c$:

\begin{equation}
    M_i = \sum_{j=1}^{m_i} \mathbf{x}_{i,j} \bigl(\omega(\bar d_{i,j}) \odot \omega(\bar d_{i,j}) \odot \bar d_{i,j}\bigr)
\end{equation}

\noindent where $\mathbf{x}_{i,j} \in \mathbb{R}^{d+1}$ denotes the $j$-th sample's bias-augmented feature vector, $\bar{d}_{i,j} \in \mathbb{R}^{1 \times |\mathcal{C}_{1:t}|}$ is the $j$-th row of $\bar{D}_i$, and $\odot$ denotes element-wise multiplication applied across classes. Geometrically, each column of $M_i$ accumulates class-specific evidence required to estimate the corresponding output neuron without storing raw samples. In the class-wise derivations that follow, $M_i$ refers to the column $M_i^{(c)}$ of the class under consideration.

$M_i$ is additive over independent data partitions, so if an additional batch yields $M'_i$, the combined moment is simply $M_i + M'_i$. This additivity is what turns federated classifier learning into a sufficient-statistics problem, because clients only need to exchange compact statistics from which the analytical solution of the pooled problem can be reconstructed exactly. The triplet $(U_i, S_i, M_i)$ thus forms the complete set of local sufficient statistics.

An important incremental learning challenge remains, however. Newly introduced output neurons are estimated exclusively from current-task data and therefore receive no negative evidence from previously learned classes $\mathcal{C}_{1:t-1}$. Without observing old-class embeddings, these neurons cannot infer which feature regions should yield low activations, often leading to over-activation on old classes and disturbing established boundaries. Unlike gradient-based classifiers, these neurons require no explicit initialization; their weights are obtained directly from the RRLS closed-form solution computed from the accumulated sufficient statistics. This limitation stems not from the RRLS solver, which solves the stated objective exactly, but from the absence of historical data in that objective. It is therefore a problem distinct from the optimization question addressed above, and it motivates the coordinated replay mechanism described next.

\subsection{Federated Coordinated Aggregation}
\label{subsec:aggregation}

To supply this missing historical context, a problem separate from the optimization question addressed by the RRLS aggregation above, each client maintains a latent replay buffer $\mathcal{B}_i$ containing embeddings from previously learned classes. Storing latent representations instead of raw images substantially reduces memory consumption while avoiding redundant backbone forward passes during replay \cite{chaudhry2019continual,bang2021rainbow}. For example, for a ResNet-18 backbone ($d=512$), each embedding requires approximately 2~KB compared with roughly 150~KB for a $224 \times 224$ RGB image, a $75\times$ memory reduction that enables scalable replay under strict storage and communication constraints.

Replay is used exclusively to calibrate newly introduced output neurons associated with $\mathcal{C}_t$. Previously learned neurons do not need it since their accumulated statistics already encode historical data, and the recursive update adds the current-task samples to them as negatives.

A second challenge arises from the federated setting itself. Because replay buffers are intentionally small, historical embeddings are greatly outnumbered by samples from the current task $\mathcal{D}_t$. Simply oversampling each client's local buffer is insufficient under Non-IID data distributions, as locally balanced replay does not guarantee a globally balanced class distribution after aggregation. Replay must therefore be coordinated across clients prior to computing local RRLS statistics.

To address this, we introduce \emph{Coordinated Replay}, a lightweight two-phase synchronization protocol executed once per incremental task (see \Cref{fig:architecture}, part B), whose objective is to align the statistical contribution of historical classes across the federation while preserving the exact analytical formulation of RRLS. This is done in two phases.

\paragraph{Phase I. Global Statistics and Quota Allocation}

Before communicating local RRLS statistics, clients exchange a compact summary of local data with the coordinator:

\begin{itemize}
    \item \textit{Reporting.} Client $i$ transmits its newly observed class tags $\mathcal{C}_{i, t}$, sample counts $n_{i, c}$ for each new class, and the per-class buffer sizes $|\mathcal{B}_{i,c}|$, i.e., the number of stored embeddings in $\mathcal{B}_i$ belonging to each previously learned class $c$.

    \item \textit{Global statistics.} The coordinator constructs the global class set $\mathcal{C}_{\mathrm{g}} = \bigcup_i \mathcal{C}_{i, t}$ and determines the maximum global class frequency $n_{\max} = \max_c (\sum_i n_{i, c})$.

    \item \textit{Quota allocation.} The coordinator assigns each client a replay quota $q_{i, c} = \left\lceil n_{\max} \frac{|\mathcal{B}_{i, c}|}{\sum_{l=1}^{N} |\mathcal{B}_{l, c}|} \right\rceil$, distributing the replay burden proportionally to the historical samples available at each client. Both $\mathcal{C}_g$ and the quotas $\{q_{i,c}\}$ are then broadcast to all clients.
\end{itemize}

\paragraph{Phase II. Coordinated Replay and Global Fusion}
\label{par:phaseII}

After receiving this information, each client constructs an augmented training set for the new neurons by oversampling embeddings from the replay buffer $\mathcal{B}_i$ according to the assigned quotas. The buffer itself remains unchanged; oversampling only provides additional evidence for recomputing the sufficient statistics $(U_i, S_i, M_i)$ which are then transmitted to the coordinator. Each client subsequently updates $\mathcal{B}_i$ with a reservoir sample of the current-task embeddings.

Because RRLS statistics are additive for moment matrices and compositional for weighted feature subspaces, the coordinator reconstructs the centralized statistics exactly, without accessing raw client data:
\begin{equation}
\begin{gathered}
M_g = \sum_{i=1}^N M_i, \\
[U_g, S_g, \sim] = \mathrm{SVD}\left([U_1S_1 \parallel U_2S_2 \parallel \cdots \parallel U_NS_N]
\right)
\end{gathered}
\end{equation}

The global classifier weights $W_g$ are recovered through the closed-form RRLS solution:

\begin{equation}
    W_g = U_g(S_g^2 + \lambda I)^{-1} U_g^\top M_g
\end{equation}

\noindent where $\lambda > 0$ is the regularization parameter. Each singular direction contributes to the weight estimate in proportion to $s^2/(s^2 + \lambda)$. Directions with large singular values are preserved nearly intact, while low-variance directions are attenuated. The parameter $\lambda$ controls numerical stability rather than a traditional bias-variance trade-off.

\subsubsection{Protocol Agnosticism and Exactness}

The aggregation above is exact. Since $M_g$ is a sum of client moments and $(U_g, S_g)$ reconstruct the subspace of the concatenated client blocks, the coordinator recovers the classifier that would be obtained by pooling the clients' data, regardless of how the data are partitioned, of the order in which the statistics arrive, and of the communication topology. Theorem~\ref{theo:equivalence} states this property formally, together with the conditions under which it holds.

\begin{theorem}[Centralized Equivalence]
Let $\{(X_i,Y_i)\}_{i=1}^N$ denote the local datasets of $N$ clients with weighted feature blocks $H_i = X_i^\top \mathbf{G}_i$, where $\mathbf{G}_i$ is the class-specific weighting matrix of Eq.~(\ref{eq:G}). Under exact arithmetic and without SVD rank truncation, the global classifier $W_g$ produced by Algorithm~\ref{alg:aggregation} satisfies $W_g = W_{\mathrm{batch}}$, where $W_{\mathrm{batch}}$ is the RRLS solution computed over the centralized feature matrix $H_{\mathrm{all}} = [H_1 \parallel H_2 \parallel \cdots \parallel H_N]$. This equivalence holds regardless of client data distribution, aggregation order, or communication topology. \emph{(Proof in \ref{apx:equivalence})}
\label{theo:equivalence}
\end{theorem}

Theorem~\ref{theo:equivalence} answers the question posed in the Introduction. Here, $W_{\mathrm{batch}}$ is computed over the same client-side training data supplied to FedIAL by the participating clients, including any replay-augmented historical evidence. The equivalence therefore concerns classifier learning over the available fixed feature space; representation adaptation remains a separate optimization problem.

\subsection{Proposed Algorithm and Complexity Analysis}
The complete federated procedure is detailed in Algorithm~\ref{alg:aggregation}, where, as in \Cref{subsec:local}, all per-class statistics are computed and aggregated class by class.

\begin{algorithm}[t]
\caption{Federated Coordinated Aggregation}
\label{alg:aggregation}
\begin{algorithmic}[1]
\Require Clients $ i = 1 \dots N$, Task $t$, Replay Buffers $\{\mathcal{B}_i\}$, Regularization $\lambda$
\State \textbf{Phase I: Statistical Alignment}
\For{each client $i$ \textbf{in parallel}}
    \State $\mathcal{C}_{i,t}, \{n_{i,c}\}, \{|\mathcal{B}_{i,c}|\} \gets \mathrm{LocalMetadata}(i)$
    \State \textbf{Send} metadata to Coordinator.
\EndFor
\State \textbf{Coordinator:} Compute $\mathcal{C}_{g}, n_{\max}, \{q_{i,c}\}$ and \textbf{Broadcast}
\State
\State \textbf{Phase II: Local Training and Global Fusion}
\For{each client $i$ \textbf{in parallel}}
    \State Oversample $\mathcal{B}_i$ based on $q_{i,c}$ to calibrate new neurons
    \State Compute local statistics $(U_i, S_i, M_i)$ via RRLS
    \State \textbf{Send} $(U_i, S_i, M_i)$ to Coordinator
\EndFor
\State \textbf{Coordinator:}
\State $\quad M_g \gets \sum_{i=1}^N M_i$
\State $\quad [U_g, S_g, \sim] \gets \mathrm{SVD}([U_1 S_1 \parallel \dots \parallel U_N S_N])$
\State $\quad W_g \gets U_g (S_g^2 + \lambda I)^{-1} U_g^\top M_g$
\State $\quad$ \textbf{Broadcast} $W_g$ to all clients
\end{algorithmic}
\end{algorithm}

\paragraph{Computational and Communication Complexity}

Locally, each client computes, for every class, one reduced SVD of the $(d+1) \times m_i$ weighted design matrix, at a cost of $\mathcal{O}\bigl((d+1)\, m_i \min(d+1, m_i)\bigr)$, together with the corresponding moment column, at a cost of $\mathcal{O}\bigl((d+1)\, m_i\bigr)$. Both are computed once per task and require no iterative optimization. Each client then transmits, per class, $U_i$, $S_i$ and the moment column, that is, $(d+1)\, r_i + r_i + (d+1)$ values, where $r_i \le \min(d+1, m_i)$ denotes the local rank. At the coordinator, global aggregation performs an SVD on the concatenated matrix of $N$ client blocks $[U_1 S_1 \parallel \cdots \parallel U_N S_N]$, which has dimension $(d+1) \times \sum_i r_i$. Using reduced SVD, global aggregation requires $\mathcal{O}\bigl((d+1)(\sum_i r_i)^2\bigr)$ operations per class, and scales linearly with the total number of classes. When clients hold few samples per class, so that $\sum_i r_i < d+1$, global aggregation is substantially less expensive than a full $(d+1) \times (d+1)$ matrix decomposition.

\section{Experiments}
\label{sec:experiments}

Our experiments are designed to answer the question posed in the Introduction along three axes: (i) whether replacing iterative federated classifier optimization with sufficient-statistic aggregation maintains or improves predictive performance and stability (\Cref{subsec:performance}--\Cref{subsec:scalability}); (ii) what efficiency this substitution buys, presented as an accuracy-efficiency trade-off rather than a single headline number; and (iii) what happens when the fixed-representation assumption is deliberately relaxed under domain shift (\Cref{subsec:domain_shift}). We compare against state-of-the-art gradient-based methods and recent analytic baselines under identical settings, and use the ablations (\Cref{subsec:ablation}) to attribute each effect to the component that produces it, with analytic aggregation accounting for the optimization cost and coordinated replay for the bias of new classes.

\subsection{Experimental Setup}

\paragraph{Datasets and Protocol}
We use CIFAR-100 \cite{krizhevsky2009learning} and TinyImageNet \cite{le2015tiny}. Unless specified, we use standard $T=10$ disjoint-class tasks. We simulate a federation of $N=30$ total clients, with $K = 10$ clients selected per training round. At each task, the protocol of \Cref{sec:method} runs over the $K$ selected clients, so the sums over $i$ in that section range over the participating clients. Results are averaged over 3 runs with seeds $\{42, 1983, 2020\}$.

\paragraph{Client Partitioning}
We evaluate both IID and heterogeneous regimes. For Non-IID, we adopt a two-stage Partial-class partition: each client observes a random fraction $\rho$ of task classes (default $\rho = 0.6$), and, within a class shared by several clients, that class's samples are split across those clients according to proportions drawn from a Dirichlet distribution $\mathrm{Dir}(\alpha)$ \cite{li2022federated} (default $\alpha = 1.0$); small $\alpha$ concentrates a class's samples on a few clients, while large $\alpha$ approaches an even split.

\paragraph{Comparison Baselines and Training Hyperparameters}
We compare with federated implementations of EWC \cite{kirkpatrick2017overcoming} and LwF \cite{li2017learning}, referred to as Fed-EWC and Fed-LwF, together with TARGET \cite{zhang_target_2023}, LANDER \cite{tran_text-enhanced_2024}, and STSA \cite{guan2025stsa}. Fed-EWC and Fed-LwF apply the original regularization strategies locally at each client and aggregate model updates via FedAvg. TARGET and LANDER optimize the full network using SGD with momentum 0.9, weight decay $5 \times 10^{-4}$, an initial learning rate of $0.01$ with cosine decay, 20 communication rounds per task, and 2 local epochs per round, following their original implementations. STSA is an analytic method that relies on gradient-based optimization only in the first task. The classifier-learning component of our method is fully analytic under the fixed-representation regime evaluated on CIFAR-100 and TinyImageNet, and therefore requires no optimizer, learning rate schedule, or local epoch tuning. The sole hyperparameter of the classifier-learning stage is the regularization term $\lambda$, set to $10^{-3}$ across all experiments; sensitivity analysis in \ref{subapx:lambda} confirms stable performance across three orders of magnitude. All baselines are run using their official code and recommended hyperparameters.

FedIAL and gradient-based baselines reflect different design philosophies. The latter fine-tune the full network through iterative federated optimization, whereas FedIAL fixes the extractor and solves the classifier analytically. Their comparison illustrates a system-level trade-off between the two paradigms. The most direct test of this paper's central claim is the comparison against STSA, which shares the fixed-representation setting but retains gradient-based optimization in its initial task, and the Linear Probing ablation (\Cref{subsec:ablation}), which holds the representation fixed and isolates iterative versus analytic classifier learning.

\paragraph{Backbone, Buffer, and Fairness Constraints}
All methods use the same ResNet-18 \cite{he2016deep} encoder with identical self-supervised pretraining \cite{DBLP:journals/corr/abs-1905-00546} on YFCC-100M \cite{thomee2015new} and Instagram-1B \cite{mahajan2018exploring}, both disjoint from all downstream FedCIL benchmarks. The encoder is frozen for analytic methods but fully fine-tuned for gradient-based baselines. Each client maintains a latent replay buffer $\mathcal{B}_i$ with a fixed capacity of $B = 500$ embeddings, using class-balanced reservoir sampling: as new classes are encountered, the budget is reallocated so that each seen class $c$ occupies exactly $\lfloor B / |\mathcal{C}_{1:t}| \rfloor$ slots. Sensitivity analysis in \ref{apx:sensitivity} confirms robust performance even as buffer capacity is significantly reduced.

\paragraph{Evaluation Metrics}
We report five complementary metrics. Let $A_{t,\tau}$ denote the global test accuracy on task $\tau$ evaluated after training on task $t \geq \tau$. \textbf{Final Accuracy} ($A_T$) is the accuracy across all seen classes after the final task. \textbf{Mean Accuracy} ($\bar{A} = \frac{1}{T}\sum_{t=1}^{T} A_t$) averages the global accuracy $A_t$ over all incremental steps. \textbf{Final Forgetting} ($F_T$) quantifies the relative drop from the highest accuracy ever achieved on each task to its final value:

\begin{equation}
   F_T = \frac{1}{T-1} \sum_{\tau=1}^{T-1} \max_{t \in \{\tau, \dots, T-1\}} \bigl(A_{t,\tau} - A_{T,\tau}\bigr).
   \label{eq:forgetting}
\end{equation}

\textbf{Stability} ($S$) and \textbf{Plasticity} ($P$) decompose the stability-plasticity trade-off \cite{Gao_2025_CVPR}. $A_{t, \tau}$ denotes the accuracy on past task $\tau$ after training task $t$, and $A_{t,t}$ the accuracy on the current task immediately after its training:

\begin{equation}
    S = \frac{1}{T-1}\sum_{t=2}^{T}\frac{1}{t-1}\sum_{\tau=1}^{t-1} A_{t,\tau},
    \qquad
    P = \frac{1}{T}\sum_{t=1}^{T} A_{t,t}.
    \label{eq:stab_plast}
\end{equation}

$S$ and $F_T$ capture different aspects of retention, so a method can have higher $S$ than a competitor while also exhibiting higher $F_T$, if it reaches substantially higher per-task peaks. Efficiency is measured through wall-clock training time and energy consumption estimated via CodeCarbon \cite{benoit_courty_2024_11171501} on an NVIDIA RTX 3090 GPU running each method in isolated controlled conditions. Code will be released upon acceptance.

\subsection{Performance and Efficiency Analysis}
\label{subsec:performance}

\begin{table}[t]
\centering
\caption{Efficiency and performance comparison on CIFAR-100 and TinyImageNet under Dirichlet partition ($\alpha=1.0$). Metrics include final accuracy ($A_T$), mean accuracy ($\bar{A}$), forgetting ($F_T$), energy consumption, and training time. Best results in \textbf{bold}.}
\label{tab:efficiency_combined}
\resizebox{\linewidth}{!}{%
\begin{tabular}{l ccc c c ccc c c}
\toprule
& \multicolumn{5}{c}{\textbf{CIFAR-100}} & \multicolumn{5}{c}{\textbf{TinyImageNet}} \\
\cmidrule(lr){2-6} \cmidrule(lr){7-11}
\textbf{Method} & $A_T \uparrow $ & $\bar{A} \uparrow $ & $F_T \downarrow$ & Eng. (Wh) $\downarrow$ & Time (m) $\downarrow$ & $A_T \uparrow $ & $\bar{A} \uparrow $ & $F_T \downarrow$ & Eng. (Wh) $\downarrow$ & Time (m) $\downarrow$ \\
\midrule
Fed-EWC & 24.96 & 41.63 & 71.36 & 369 & 60 & 17.83 & 37.06 & 70.47 & 676 & 104 \\
Fed-LwF & 37.89 & 51.78 & 42.53 & 272 & 42 & 31.08 & 51.09 & 45.74 & 509 & 76 \\
TARGET & 5.20 & 31.71 & 42.18 & 1,936 & 375 & 34.18 & 53.54 & 38.25 & 16,880 & 2,558 \\
LANDER & 28.66 & 44.75 & 64.75 & 2,152 & 369 & 11.90 & 16.74 & 16.53 & 3,830 & 616 \\
STSA   & 51.95 & 62.36 & \textbf{11.62} & \textbf{95} & \textbf{25} & 39.40 & 50.06 & \textbf{6.14} & \textbf{130} & \textbf{31} \\
\midrule
\textbf{FedIAL} & \textbf{57.53} & \textbf{70.13} & 13.97 & 125 & 35 & \textbf{53.25} & \textbf{61.30} & 10.93 & 380 & 102 \\
\bottomrule
\end{tabular}}
\end{table}

\Cref{tab:efficiency_combined} summarizes the performance on CIFAR-100 and TinyImageNet. Fed-EWC and Fed-LwF, despite being grounded in principled regularization, suffer severe accuracy collapse in the federated setting, with catastrophic forgetting dominating ($F_T > 40\%$) because local penalty terms computed on heterogeneous data fail to protect globally shared knowledge, and gradient averaging introduces conflicting update directions. TARGET and LANDER fare no better, with TARGET collapsing to near-chance accuracy on CIFAR-100 under Non-IID conditions. FedIAL outperforms all gradient-based methods and surpasses the analytic competitor STSA, reaching \textbf{57.53\%} $A_T$ on CIFAR-100 (+5.6 points over STSA) and \textbf{53.25\%} on TinyImageNet (+13.8 points). Under heterogeneous federation, replacing iterative classifier optimization with sufficient-statistic aggregation, combined with coordinated replay for the new-class evidence problem, is thus more effective than both classical regularization and generative replay, while requiring a single synchronization cycle per task.

\paragraph{Efficiency Trade-off}
The efficiency results are best read as an accuracy-efficiency trade-off rather than a single maximum-reduction figure. Under the evaluated settings, FedIAL is roughly $10\times$ faster and consumes over 90\% less energy than the generative-replay baselines TARGET and LANDER, the most expensive methods in the comparison. Against the regularization baselines Fed-EWC and Fed-LwF, FedIAL still consumes less energy on both benchmarks (25--66\% less), but runtimes are comparable, and Fed-LwF is faster on TinyImageNet. These savings follow from the formulation, since with iterative classifier optimization removed, each task requires a single forward pass per sample, a closed-form solve, and one synchronization cycle. Within the analytic regime, the trade-off runs the other way. STSA reports the lowest training time and energy, and FedIAL incurs additional cost (1.3$\times$ the energy of STSA on CIFAR-100 and about 3$\times$ on TinyImageNet) in exchange for accuracy gains of +5.6 and +13.8 points, respectively. This overhead comes from the per-class formulation. Unlike STSA, which maintains a single shared classifier, FedIAL maintains independent sufficient statistics per output neuron, each class having its own $(U^{(c)}, S^{(c)}, M^{(c)})$, so that every class boundary is solved in closed form on its own, at a per-task SVD and storage cost that grows linearly with the number of classes.

\begin{figure}[t]
    \centering
    \includegraphics[width=0.8\linewidth]{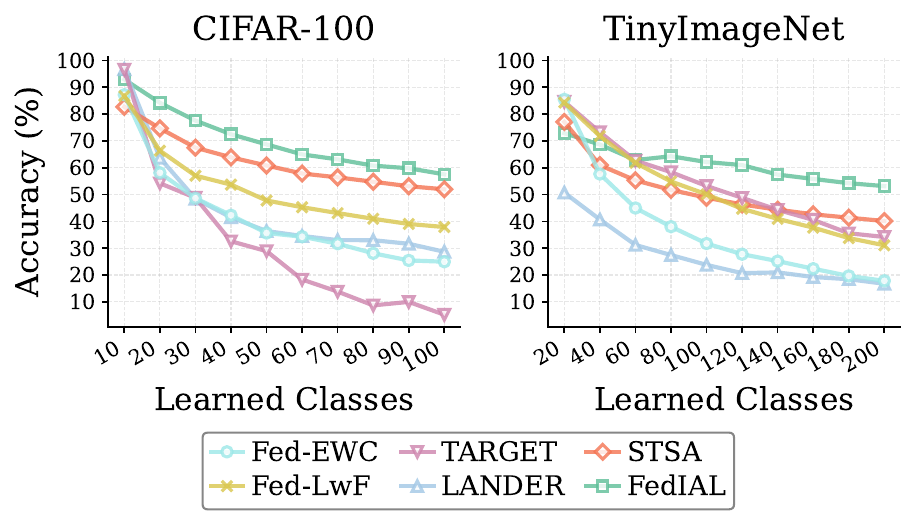}
    \caption{Accuracy evolution across 10 sequential tasks.}
    \label{fig:curves_accuracy}
\end{figure}

The evolution of test accuracy across the sequential arrival of 10 tasks is visualized in \Cref{fig:curves_accuracy}. Gradient-based methods (TARGET, LANDER) suffer sharp post-task degradation from catastrophic forgetting; our method decays gracefully. On TinyImageNet, our approach maintains a significant margin over STSA throughout the incremental process (task $t > 1$), suggesting that our analytical formulation preserves the feature manifold's discriminative power as the global label space expands.

\begin{figure}[t]
    \centering
    \includegraphics[width=0.75\linewidth]{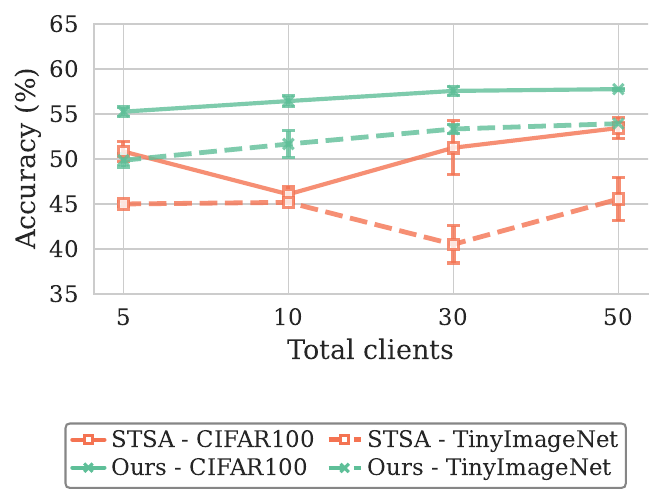}
    \caption{Final accuracy ($A_T$) as a function of the total client pool size $N$ on CIFAR-100 and TinyImageNet, with the selection budget fixed to $K = \min(10, N)$. Error bars denote standard deviation over 3 runs.}
    \label{fig:clients}
\end{figure}

We further investigate the impact of the client pool size $N$ in \Cref{fig:clients}, varying $N$ from 5 to 50 while fixing $K = \min(10, N)$. A larger pool increases system heterogeneity, since data becomes sparser across the federation. Despite this, our analytical approach is remarkably stable across pool sizes on both datasets; a plausible explanation is that a larger pool provides more comprehensive coverage of the global data distribution, whereas a 5-client configuration leads to dense but less representative local distributions that can cause diversity loss during temporal oversampling. STSA, in contrast, exhibits a less stable trend and consistently larger standard deviations, indicating higher sensitivity to the federation configuration. These results suggest that reconstructing the classifier from aggregated sufficient statistics provides more consistent performance as the number of clients increases, since the reconstructed solution does not depend on how the data happen to be partitioned.

\begin{figure}[t]
    \centering
    \includegraphics[width=0.85\linewidth]{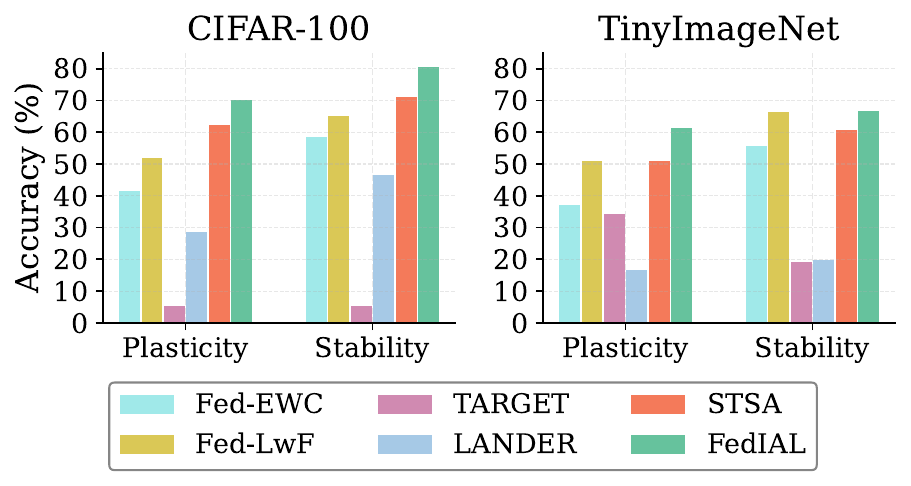}
    \caption{Plasticity-stability trade-off. Plasticity is accuracy on newly introduced classes immediately after training; Stability ($S$) is average raw accuracy retained on all prior classes. FedIAL achieves higher absolute retention than STSA on both metrics despite a larger peak-to-final gap in $F_T$.}
    \label{fig:plasticity_stability}
\end{figure}

\Cref{fig:plasticity_stability} decomposes the plasticity-stability trade-off. Gradient-based methods suffer a stability collapse, where acquiring new tasks erases previously learned knowledge (e.g., TARGET), while STSA achieves strong stability at the cost of low plasticity. FedIAL demonstrates superior retention while maintaining significantly higher plasticity than either alternative.

A clarification is warranted regarding the relationship between $S$ and $F_T$. Table~\ref{tab:efficiency_combined} shows a higher $F_T$ for FedIAL than for STSA, whereas \Cref{fig:plasticity_stability} shows FedIAL achieves higher $S$. This is not a contradiction, since FedIAL reaches substantially higher per-task peak accuracies (\Cref{fig:curves_accuracy}), it can experience a larger drop from its own peak while still retaining a higher absolute accuracy than STSA, so $S$ remains higher despite a larger forgetting rate.

\subsection{Robustness to Data Heterogeneity}

\begin{table}[t]
\caption{Robustness to data heterogeneity in CIFAR-100 across Dirichlet concentration parameters ($\alpha$), which control how evenly a shared class's samples are split across the clients that hold it. Lower $\alpha$ concentrates a class's samples on fewer clients (higher quantity skew); higher $\alpha$ approaches an even split. Best results in \textbf{bold}. }
\label{tab:alpha_comparison}
\centering
\resizebox{0.75\linewidth}{!}{%
\begin{tabular}{l cc cc cc  cc}
    \toprule
    & \multicolumn{2}{c}{IID ($\alpha = \infty$)} & \multicolumn{2}{c}{$\alpha = 1.0$} & \multicolumn{2}{c}{$\alpha = 0.5$} & \multicolumn{2}{c}{$\alpha = 0.1$} \\
    \cmidrule(r{1.5em}){2-3} \cmidrule(r{1.5em}){4-5} \cmidrule(r{1.5em}){6-7} \cmidrule{8-9}
    Method & $A_T \uparrow$ & $F_T \downarrow$ & $A_T \uparrow$ & $F_T \downarrow$ & $A_T \uparrow$ & $F_T \downarrow$ & $A_T \uparrow$ & $F_T \downarrow$ \\
    \midrule
    Fed-EWC & 25.41 & 72.72 & 24.96 & 71.36 & 23.94 & 69.79 & 21.06 & 60.70 \\
    Fed-LwF & 41.82 & 41.23 & 37.89 & 42.53 & 38.01 & 43.07 & 27.45 & 45.07 \\
    TARGET & 12.54 & 35.84 & 5.20 & 42.18 & 6.86 & 34.15 & 4.30 & 32.27\\
    LANDER & 33.40 & 59.11 & 28.66 & 64.75 & 28.27 & 64.60 & 27.72 & 65.82 \\
    STSA & 55.30 & \textbf{11.57} & 51.95 & \textbf{11.62} & 53.72 & \textbf{11.97} & 50.37 & \textbf{11.23} \\
    \midrule
    \textbf{FedIAL (Ours)} & \textbf{57.93} & 13.37 & \textbf{57.53} & 13.97 & \textbf{57.40} & 13.53 & \textbf{57.84} & 13.76\\
    \bottomrule
\end{tabular}}
\end{table}

A critical failure scenario for FedCIL, inherited from FL, is representation drift under Non-IID data. \Cref{tab:alpha_comparison} evaluates robustness by varying $\alpha$ from IID to severe sample-level skew ($\alpha=0.1$). Gradient-based methods remain well below the analytic ones at every level. TARGET effectively collapses already at moderate heterogeneity (from 12.54 to 5.20 $A_T$ at $\alpha = 1.0$), and Fed-EWC remains low across all levels (from 25.41 to 21.06), as locally computed regularization penalties diverge across clients and gradient averaging amplifies rather than corrects the imbalance. Fed-LwF loses over 14 points of $A_T$ at $\alpha=0.1$, and STSA drops $\approx 5$ points. FedIAL holds steady at $\sim57\%$ across all heterogeneity levels, including $\alpha = 0.1$. This invariance is consistent with Theorem~\ref{theo:equivalence}, under whose assumptions the aggregated RRLS classifier is equivalent to the centralized solution regardless of client partitioning, so heterogeneity affects only how the same sufficient statistics are distributed across clients, not the solution reconstructed from them. The small residual variations across $\alpha$ (below 0.6 points) stem from the replay buffers rather than from the solver. Since class-balanced reservoir sampling selects a random subset of embeddings at each client, the replay-augmented data supplied to RRLS differ slightly from one partition to another, and the equivalence holds for the data actually supplied, not for the samples that reservoir sampling would have selected under a different partition.

\subsection{Scalability to Realistic Incremental Steps}
\label{subsec:scalability}

\begin{table}[t]
\caption{Performance comparison in many-task scenarios. We report final accuracy ($A_T$), mean accuracy ($\bar{A}$), and forgetting ($F_T$) in \%. Best results are \textbf{bolded}. Gradient-based baselines were omitted due to prohibitive runtime ($>10\times$ wall-clock time of analytic methods) in the 40--50 task regime.}
\label{tab:real_setting_small_tasks}
\centering
\begin{tabular}{lccc ccc}
\toprule
& \multicolumn{3}{c}{CIFAR-100 (50 tasks)}
& \multicolumn{3}{c}{TinyImageNet (40 tasks)} \\
\cmidrule{2-7}
Method &
$A_T \uparrow$ & $\bar{A} \uparrow$ & $F_T \downarrow$ &
$A_T \uparrow$ & $\bar{A} \uparrow$ & $F_T \downarrow$ \\
\midrule
STSA & 51.80 & 61.79 & 14.53 & 48.34 & 59.77 & \textbf{12.00} \\
\textbf{FedIAL} & \textbf{57.57} & \textbf{71.81} & \textbf{13.77} & \textbf{53.83} & \textbf{62.74} & 12.52 \\
\bottomrule
\end{tabular}
\end{table}

Real-world deployments often involve streaming tasks with few classes rather than large batched tasks. We evaluate this ``Real Problem Setting'' in \Cref{tab:real_setting_small_tasks} with 50 tasks (CIFAR-100) and 40 tasks (TinyImageNet). FedIAL retains its advantage over STSA in this regime. On CIFAR-100, the gap is preserved (+5.8 points in $A_T$, compared with +5.6 in the 10-task setting) and FedIAL, notably, achieves \emph{lower} forgetting than STSA (13.77\% vs.\ 14.53\%). On TinyImageNet, the gap narrows from +13.8 to +5.5 points, mainly because STSA benefits from the finer task granularity (from 39.40 to 48.34 $A_T$), whereas FedIAL remains essentially unchanged (53.83 vs.\ 53.25). Gradient baselines are omitted due to prohibitive runtime; as shown in the standard setting, they also consistently underperform analytic approaches in both accuracy and forgetting.

\subsection{Controlled Relaxation of the Fixed-Representation Assumption via Lightweight Adaptation}
\label{subsec:domain_shift}

The central claim of this paper concerns classifier learning \emph{after} a representation has been fixed; it presumes that the pretrained backbone yields sufficiently transferable features. This holds for benchmarks such as CIFAR-100 and TinyImageNet, but real-world target distributions may diverge substantially from the backbone's pretraining domain, rendering frozen features less discriminative. Here we study a controlled relaxation of the fixed-representation assumption, in which a one-time adaptation stage is allowed before the incremental phase begins, after which the same analytical classifier-learning formulation applies unchanged. The experiment is not intended to show that the entire FedCIL pipeline is always analytic, since representation adaptation, where needed, remains an iterative gradient-based process. Its purpose is to delimit where the analytical claim applies and how little adaptation is required to restore its premise.

We evaluate FedIAL on ImageNet-R \cite{hendrycks2021many} (200 ImageNet classes rendered in diverse artistic styles) using a ViT-B/16 \cite{dosovitskiy2020image} backbone pretrained on ImageNet-21K. Rather than abandoning the frozen-backbone paradigm entirely, we introduce a lightweight adaptation stage exclusively for the first task using trainable bottleneck adapters \cite{houlsby2019parameter}. Inserted in parallel with each transformer block's feed-forward sublayer, these adapters are optimized via FedAvg during task 1 alongside a temporary cross-entropy head, while all original backbone parameters remain frozen. Subsequently, the adapters are frozen and all remaining tasks follow the standard FedIAL protocol, learning exclusively through the analytic RRLS head; that is, from task~2 onward the fixed-representation regime holds again and no iterative classifier optimization occurs. This adaptation stage relies on iterative gradient-based federated optimization of the adapter parameters (\ref{apx:adapter:protocol}); it is classifier learning across the subsequent incremental tasks, not the representation itself, that remains free of iterative optimization. Additional implementation details are provided in \ref{apx:adapter}.

Table~\ref{tab:imagenetr} reports results obtained using the same pretrained ViT-B/16 backbone for all compared methods, including STSA, together with two ablations of FedIAL: (i) a fully frozen backbone without adaptation, and (ii) adapters trained throughout all incremental tasks. Gradient-based baselines \cite{zhang_target_2023, tran_text-enhanced_2024} are omitted due to the prohibitive cost of repeatedly fine-tuning the extractor in a federated setting; we similarly exclude ViT-native FedCIL methods such as PILoRA \cite{guo2024pilora} and FCILPT \cite{luo2025federated}, since a meaningful comparison would require reproducing their dedicated training protocols and hyperparameter selections, an effort orthogonal to this section's goal of validating the fixed-representation relaxation, which we leave to future work.

\begin{table}[t]
\caption{Performance on ImageNet-R (200 classes, 10 tasks) with ViT-B/16. \emph{w/o adapt.} denotes a fully frozen backbone; \emph{all tasks} trains adapters at every task increment. Best results in \textbf{bold}.}
\label{tab:imagenetr}
\centering
\begin{tabular}{lccc cc}
    \toprule
    Method & $A_T \uparrow$ & $\bar{A} \uparrow$ & $F_T \downarrow$
           & Eng.\ (Wh) $\downarrow$ & Time (m) $\downarrow$ \\
    \midrule
    STSA \cite{guan2025stsa}      & 62.97 & 70.35 & 6.48 & \textbf{401} & \textbf{94} \\
    \midrule
    FedIAL w/o adapt.             & 66.04 & 74.06 & 3.82 & 504 & 122 \\
    FedIAL (all tasks)            & \textbf{72.20} & \textbf{81.81} & 3.93 & 2,400 & 480 \\
    \midrule
    \textbf{FedIAL (task 1 only)} & 72.00 & 80.05 & \textbf{2.97}  & 588 & 142 \\
    \bottomrule
\end{tabular}
\end{table}

A fully frozen ViT-B/16 already surpasses STSA (66.04 vs.\ 62.97 $A_T$), indicating that the pretrained representation retains substantial transferability despite the domain shift; the assumption degrades gracefully rather than failing outright. Adapting solely on task~1 boosts performance to 72.00 $A_T$ while achieving the lowest forgetting among all compared configurations (2.97), so in this benchmark a single, bounded adaptation phase suffices to restore the premise of the analytical formulation. Adapting across all tasks, that is, abandoning the fixed representation entirely, yields only a marginal accuracy gain (72.20 $A_T$) but worsens forgetting (3.93), quadruples energy consumption (2,400 vs.\ 588 Wh), and proportionally increases training time. Under this benchmark, nearly all of the benefit of representation adaptation is obtained by adapting once and freezing, after which incremental classifier learning proceeds analytically at low cost, again following the accuracy-efficiency behavior established earlier in this section. This supports the scoping of the analytical claim, in which iterative optimization, when needed at all, is confined to an occasional, bounded representation-alignment step and never enters incremental classifier learning.

\subsection{Ablation Study}
\label{subsec:ablation}

The ablation in \Cref{tab:ablation} separates the paper's two components along the two problems they were introduced to solve. The analytic solver and sufficient-statistic aggregation address the optimization question (removing iterative classifier optimization and its cost), while coordinated replay addresses the incremental bias of newly introduced classes.

\textbf{Coordination vs.\ Local Replay (the missing-evidence problem).} Replacing Phase I with \textit{Local-Only} oversampling leads to a sharp collapse (72.20\% forgetting on CIFAR-100). Without global metadata exchange, local balancing introduces conflicting class biases that prevent the formation of a stable global discriminative boundary. This isolates the contribution of coordinated replay. It does not change the optimization, since the RRLS solve is identical in both configurations, but it supplies and balances the historical negative evidence that new output neurons otherwise lack under Non-IID FedCIL.

\textbf{Analytic vs.\ Iterative Optimization (the optimization question).} We compare FedIAL against an iterative \textit{Linear Probing} baseline using SGD over the same frozen features. Despite its higher computational budget (multiple rounds and local epochs), iterative probing achieves significantly lower accuracy. Because the representation is identical in both configurations, this comparison isolates the optimization strategy itself. Even over frozen features, iteratively optimized and averaged classifiers drift under Non-IID data, whereas the analytic approach reconstructs the solution of the centralized RRLS objective in a single synchronization cycle. The efficiency gains reported in \Cref{subsec:performance} originate from this component, which also contributes to retention, since iterative probing more than doubles forgetting with respect to the analytic solver (33.37\% vs.\ 13.97\% on CIFAR-100). Forgetting reduction is therefore shared between both components, with coordinated replay accounting for the largest share, as its removal raises $F_T$ to 72.20\%.

\begin{table}[t]
\caption{Ablation study on CIFAR-100 and TinyImageNet (10-task split). \textit{Local-Only} refers to removing Phase I coordination. \textit{Linear Probing} replaces the RRLS solver with iterative SGD.}
\label{tab:ablation}
\centering
\begin{tabular}{lccc ccc}
\toprule
& \multicolumn{3}{c}{CIFAR-100} & \multicolumn{3}{c}{TinyImageNet} \\
\cmidrule{2-4} \cmidrule{5-7}
Configuration & $A_T \uparrow$ & $\bar{A} \uparrow$ & $F_T \downarrow$ & $A_T \uparrow$ & $\bar{A} \uparrow$ & $F_T \downarrow$ \\
\midrule
Local-Only & 24.16 & 54.06 & 72.20 & 21.05 & 48.12 & 68.45 \\
Linear Probing (SGD) & 48.67 & 57.14 & 33.37 & 45.83 & 53.57 & 26.31 \\
\textbf{FedIAL (Full)} & \textbf{57.53} & \textbf{70.13} & \textbf{13.97} & \textbf{53.25} & \textbf{61.30} & \textbf{10.93} \\
\bottomrule
\end{tabular}
\end{table}

\section{Conclusion}
\label{sec:conclusion}

This work asked whether federated class-incremental classifier learning requires iterative federated optimization once a sufficiently transferable representation is available and kept fixed. We showed, formally and empirically, that it does not. Under the fixed-representation regime, classifier learning is a convex RRLS problem determined by additive sufficient statistics, so the server can reconstruct the classifier, equivalent to the centralized RRLS solution under exact arithmetic and without lossy rank truncation, from a single exchange of client statistics per task. The broader takeaway is a change of perspective: when a pretrained representation is sufficiently transferable, FedCIL classifier learning can be treated as a \emph{distributed sufficient-statistics problem} rather than an iterative optimization problem. We propose FedIAL, which operationalizes this observation and additionally resolves the separate class-incremental problem that remains, the missing historical negative evidence of newly introduced classes, through globally coordinated latent replay. Experiments confirm that this formulation maintains or improves predictive performance and stability over iterative and analytic baselines under the evaluated settings, while offering a favorable accuracy-efficiency trade-off, since iterative optimization is removed from classifier learning entirely and confined, at most, to a bounded representation-alignment step.

A natural remaining scientific question is \emph{when the fixed-representation assumption is valid}, and \emph{how much representation adaptation is actually necessary when it is not}. Our ImageNet-R study gives a first data point, showing that even under substantial domain shift a single bounded adaptation phase, with adapters trained once and then frozen, recovered nearly all of the benefit of continuous adaptation at a fraction of its cost, after which classifier learning proceeded analytically.

This question also delimits the method's scope and structures our future work. FedIAL's effectiveness depends on the quality of the pretrained backbone's representations; in domains where suitable pretrained features are unavailable or poorly transferable, such as medical imaging or satellite data, the fixed-representation assumption may limit performance. The natural research direction is therefore to characterize the boundary between three regimes, fully frozen representations, lightweight bounded adaptation, and full fine-tuning, as a function of domain shift and data scarcity, and to make the choice among them principled rather than fixed in advance. Our task-1 adapter mechanism is a first, deliberately coarse instrument for probing this boundary, since adapters are inserted uniformly across all transformer blocks and trained only once, at the very first task. A more refined mechanism could instead determine \emph{where} (which blocks) and \emph{when} (which task, and for which clients) adaptation is actually needed, and could draw on adapter-composition frameworks such as AdapterFusion \cite{pfeiffer2021adapterfusion} to reuse or combine adapters learned by different clients or tasks rather than discarding them. We leave a systematic comparison against ViT-native FedCIL baselines such as PILoRA \cite{guo2024pilora} and FCILPT \cite{luo2025federated}, and the replacement of the latent replay buffer with class-conditioned generative mechanisms \cite{van2018generative} to further reduce local storage requirements, for future work.

\section*{Declaration of competing interest}
The authors declare that they have no known competing financial interests or personal relationships that could have appeared to influence the work reported in this paper.

\section*{Data availability}
The code and datasets generated and analyzed during the current study will be made available in a public GitHub repository upon publication acceptance.

\section*{Funding}
\begin{sloppypar}
Work funded by Project PID2023-147404OB-I00 (MICIU / AEI / 10.13039 / 501100011033; ERDF/EU; ESF+/EU), Horizon Europe (GA 101070381), the Ministry for Digital Transformation and Civil Service and Next-GenerationEU/PRTR (TSI-100925-2023-1) and by the Competitive Reference Groups of the Xunta de Galicia (grant ED431C 2026/38).  CITIC, as a member of the CIGUS Network, receives subsidies from the ``Xunta de Galicia" and from the ERDF Operational Programme Galicia 2021-2027 (Grant ED431G 2023/01).
\end{sloppypar}

\section*{CRediT authorship contribution statement}

\textbf{Alejandro Dopico-Castro:} Writing -- review \& editing, Writing -- original draft, Software, Methodology, Investigation, Data curation, Conceptualization. \textbf{Oscar Fontenla-Romero:} Writing -- review \& editing, Supervision, Methodology, Investigation, Conceptualization. \textbf{Bertha Guijarro-Berdiñas:} Writing -- review \& editing, Supervision, Methodology, Investigation, Conceptualization. \textbf{Amparo Alonso-Betanzos:} Writing -- review \& editing, Supervision, Methodology, Investigation, Conceptualization.

\bibliographystyle{model1-num-names}
\bibliography{main}

\appendix

\section{Mathematical Equivalence of RRLS}
\label{apx:equivalence}

We show that the proposed federated analytic head produces a global weight matrix $W_g$ that is mathematically equivalent to the centralized solution $W_{\mathrm{batch}}$ that would be obtained if all client-side data used by the protocol were available simultaneously in a single location. We assume exact arithmetic and no lossy rank truncation during SVD computation.

\begin{theorem*}[Centralized Equivalence, restated]
Under exact arithmetic and without lossy rank truncation, the global weights $W_g$ obtained via the federated statistical aggregation protocol (Algorithm~\ref{alg:aggregation}) are mathematically equivalent to the batch Regularized Recursive Least Squares (RRLS) solution $W_\mathrm{batch}$ computed on the centralized dataset, regardless of client data partitioning, non-IID heterogeneity, or communication topology.
\end{theorem*}

\begin{proof}
The proof proceeds in four steps: formulating the centralized solver, demonstrating additivity of the moment matrix, proving equivalence of the SVD fusion, and establishing invariance to the aggregation topology.

\paragraph{Step 1: The Centralized Solver}
Let each client $i \in \{1, \dots, N\}$ have a local feature matrix $X_i \in \mathbb{R}^{m_i \times (d+1)}$ (embeddings augmented with a bias column) and a diagonal sample-weighting matrix $\mathbf{G}_i \in \mathbb{R}^{m_i \times m_i}$, defined in Eq.~(\ref{eq:G}). As in \Cref{subsec:local}, the class index is omitted: $(U, S)$ and the RRLS solution refer to a single output class, and $M$ to the corresponding column of the moment matrix. Define the weighted feature blocks for client $i$ as:
\begin{equation}
    H_i = X_i^\top \mathbf{G}_i \in \mathbb{R}^{(d + 1) \times m_i}
\end{equation}

The centralized feature matrix is the horizontal concatenation of all client blocks:
\begin{equation}
    H_{\mathrm{all}} = [H_1 \parallel H_2 \parallel \dots \parallel H_N ] \in \mathbb{R}^{(d+1) \times m} \quad m = \textstyle\sum_i m_i.
\end{equation}

Its reduced SVD gives $[U_{\mathrm{all}}, S_{\mathrm{all}}, \sim] = \mathrm{SVD}(H_{\mathrm{all}})$, with $U_{\mathrm{all}} \in \mathbb{R}^{(d+1) \times r}$, $S_{\mathrm{all}} \in \mathbb{R}^{r \times r}$, $r = \min(d+1,m)$. Given the centralized moment matrix $M_{\mathrm{all}} \in \mathbb{R}^{(d+1)\times|\mathcal{C}_{1:t}|}$, the batch RRLS solution is:

\begin{equation}
    W_{\mathrm{batch}} = U_{\mathrm{all}} (S_{\mathrm{all}}^2 + \lambda I)^{-1} U_{\mathrm{all}}^\top M_{\mathrm{all}}
\end{equation}

\noindent where $\lambda > 0$ is the regularization parameter.

\paragraph{Step 2: Additivity of the Moment Matrix}
Each client computes its local moment matrix $M_i$ independently over its local dataset $\mathcal{D}_i$:

\begin{equation}
    M_i = \sum_{j=1}^{m_i} \mathbf{x}_{i,j} \, g_{i,j}^\top \in \mathbb{R}^{(d+1) \times |\mathcal{C}_{1:t}|}
\end{equation}

\noindent where $\mathbf{x}_{i,j} \in \mathbb{R}^{d+1}$ is the bias-augmented embedding of the $j$-th local sample and $g_{i,j} = (\omega (\bar{d}_{i,j}) \odot \omega (\bar{d}_{i,j}) \odot \bar{d}_{i,j}) \in \mathbb{R}^{|\mathcal{C}_{1:t}|}$ depends only on that sample. The global moment matrix is defined as the sum over all clients:

\begin{equation}
\begin{aligned}
    M_g
    &= \sum_{i=1}^{N} M_i \\
    &= \sum_{i=1}^{N} \left(\sum_{j=1}^{m_i} \mathbf{x}_{i,j}\, g_{i,j}^\top \right) \\
    &= M_{\mathrm{all}} .
\end{aligned}
\end{equation}

Since matrix addition is commutative and associative, the order in which clients contribute does not affect the result.

\paragraph{Step 3: Equivalence of SVD Fusion}

Let $[U_i, S_i, \sim] = \mathrm{SVD}(H_i)$ be the local reduced SVD, so $H_i H_i^\top = U_i S_i^2 U_i^\top$. Because $H_{\mathrm{all}} H_{\mathrm{all}}^\top = \sum_{i=1}^N H_i H_i^\top = \sum_{i=1}^N U_i S_i^2 U_i^\top$, the left singular subspace of $H_{\mathrm{all}}$ is the same as that of the horizontally concatenated scaled-basis matrix $[U_1 S_1 \parallel \cdots \parallel U_N S_N]$, since:

\begin{equation}
\begin{aligned}
    &[U_1 S_1 \parallel \cdots \parallel U_N S_N]
    \left([U_1 S_1 \parallel \cdots \parallel U_N S_N]\right)^\top \\
    &= \sum_{i=1}^N U_i S_i^2 U_i^\top \\
    &= H_{\mathrm{all}} H_{\mathrm{all}}^\top .
\end{aligned}
\end{equation}

Therefore, computing:

\begin{equation}
    [U_g, S_g, \sim] = \mathrm{SVD} \left([U_1 S_1 \parallel U_2 S_2 \parallel \cdots \parallel U_N S_N]\right)
\end{equation}

yields $U_g S_g^2 U_g^\top = U_{\mathrm{all}} S_{\mathrm{all}}^2 U_{\mathrm{all}}^\top$, and hence $U_g U_g^\top = U_{\mathrm{all}} U_{\mathrm{all}}^\top$. The equality holds exactly; individual columns of $U_g$ may differ from those of $U_{\mathrm{all}}$ by sign or ordering, but these ambiguities cancel in the projection $U U^\top$ and do not affect $W_g$. The same argument applies inductively: if $U_{\mathrm{prev}}, S_{\mathrm{prev}}$ represent the accumulated subspace of clients $1, \dots, i$, then

\begin{equation}
    [U_{\mathrm{new}}, S_{\mathrm{new}}, \sim] = \mathrm{SVD} \left([U_{\mathrm{prev}} S_{\mathrm{prev}} \parallel U_{i+1} S_{i+1}]\right)
\end{equation}

exactly reconstructs the subspace of clients $1, \dots, i+1$ by the same identity above, completing the induction.

\paragraph{Step 4: Invariance to Aggregation Topology}

Whether statistics are collected via a Star topology (simultaneous aggregation) or a Ring topology (sequential node-by-node fusion), the results are identical. For the moment matrix, associativity of addition gives:

\begin{equation}
\begin{aligned}
    M_g^{\mathrm{ring}}
    &= \bigl(\cdots (M_1 + M_2) + \cdots + M_N \bigr) \\
    &= \sum_{i=1}^N M_i \\
    &= M_{\mathrm{all}} .
\end{aligned}
\end{equation}

For the SVD fusion, sequential applications of the incremental update in Step~3 yield the same $U_g U_g^\top$ regardless of processing order, since each step reconstructs the exact subspace of the accumulated blocks.

\paragraph{Conclusion}

Since $U_g U_g^\top = U_{\mathrm{all}} U_{\mathrm{all}}^\top$, $S_g = S_{\mathrm{all}}$, and $M_g = M_{\mathrm{all}}$, and since both $S_g$ and $S_\mathrm{all}$ are the square roots of the eigenvalues of the same matrix $H_{\mathrm{all}}H_{\mathrm{all}}^\top$ (which are unique), the shared eigendecomposition ensures $U_g(S_g^2 + \lambda I)^{-1}U_g^\top = U_{\mathrm{all}} (S_{\mathrm{all}}^2+\lambda I)^{-1} U_{\mathrm{all}}^\top$. Substituting the federated statistics into the global solver therefore gives:

\begin{equation}
\begin{aligned}
    W_g
    &= U_g (S_g^2 + \lambda I)^{-1} U_g^\top M_g \\
    &= U_{\mathrm{all}}(S_{\mathrm{all}}^2+\lambda I)^{-1}
       U_{\mathrm{all}}^\top M_{\mathrm{all}} \\
    &= W_{\mathrm{batch}} .
\end{aligned}
\end{equation}

Therefore $W_g$ is the unique minimizer of the centralized RRLS objective, invariant to client partitioning, aggregation order, and communication topology.
\end{proof}

\section{Sensitivity Analysis}
\label{apx:sensitivity}

\subsection{Regularization Parameter $\lambda$}
\label{subapx:lambda}
The only hyperparameter of the classifier-learning stage is the regularization term $\lambda$ in the closed-form solution. Unlike gradient-based methods, which require careful tuning of learning rates, momentum, and scheduling, our analytic solver is remarkably insensitive to this parameter. \Cref{tab:lambda_sensitivity} reports final accuracy across three orders of magnitude, from $\lambda = 10^{-3}$ to $\lambda = 1$. Performance remains stable at $57.53\%$ throughout this range, confirming that the RRLS objective is well-conditioned and that our default choice of $\lambda = 10^{-3}$ is a conservative, principled default rather than a carefully optimized selection. This robustness follows from the analytic nature of the solver. Since no iterative optimization is performed for classifier learning, there is no risk of divergence or oscillation due to poor hyperparameter choices, and the regularizer serves primarily to ensure numerical invertibility rather than to control a bias-variance trade-off in the classical sense.

\begin{table}[h]
\caption{Sensitivity to regularization weight $\lambda$. Final accuracy remains stable across a wide range of $\lambda$ values.}
\label{tab:lambda_sensitivity}
\centering
\begin{tabular}{lc}
\toprule
    $\lambda$ & Accuracy ($\%$) $\uparrow$ \\
\midrule
    0.001 & 57.53 \\
    0.01  & 57.53 \\
    0.1   & 57.53 \\
    1     & 57.53 \\
\bottomrule
\end{tabular}
\end{table}

\subsection{Replay Buffer Capacity}
While our method is hyperparameter-free with respect to classifier optimization, the replay buffer size $B$ governs the fidelity of historical class representations retained per client. Table~\ref{tab:buffer_sensitivity} reports final accuracy as $B$ is varied from 100 to 2000 samples per client. Performance improves monotonically with buffer capacity, from $53.12\%$ at $B{=}100$ to $57.78\%$ at $B{=}2000$, at which point the curve saturates. Even at our default capacity of $B{=}500$, the method achieves $57.53\%$, closely matching settings with four times the memory budget, attributable to the class-balanced reservoir sampling strategy and the low per-sample storage cost of latent embeddings.

\begin{table}[h]
\caption{Sensitivity to replay buffer size $B$ on CIFAR-100. Performance degrades gracefully with smaller buffers, remaining competitive at moderate capacities.}
\label{tab:buffer_sensitivity}
\centering
\begin{tabular}{lc}
\toprule
    Buffer Size $B$ & Accuracy ($\%$) $\uparrow$ \\
\midrule
    100  & 53.12 \\
    200  & 54.84 \\
    500  & 57.53 \\
    1000 & 57.61 \\
    2000 & 57.78 \\
\bottomrule
\end{tabular}
\end{table}

\section{Lightweight Adapter Extension: Architecture and Implementation}
\label{apx:adapter}

This appendix provides the architectural specification, federated protocol, and hyperparameter details for the adapter-based relaxation of the fixed-representation assumption used in \Cref{subsec:domain_shift}.

\subsection{Adapter Architecture}
\label{apx:adapter:arch}

We insert a bottleneck adapter module in \emph{parallel} with the feed-forward network (FFN) sublayer of each ViT-B/16 transformer block. Let $\mathbf{h} \in \mathbb{R}^{d}$ denote the hidden state entering the FFN sublayer. The adapted block computes:

\begin{equation}
    \mathbf{h}' = \mathrm{FFN}(\mathbf{h}) + \gamma \mathrm{Adapter}(\mathbf{h}),
    \label{eq:adapter_parallel}
\end{equation}

\noindent where the adapter is a two-layer bottleneck projection:
\begin{equation}
    \begin{gathered}
    \operatorname{Adapter}(\mathbf{h}) = \mathbf{W}_{up} \operatorname{GELU} \bigl(\mathbf{W}_{down} \mathbf{h}\bigr), \\
    \mathbf{W}_{down} \in \mathbb{R}^{r_a \times d}, \mathbf{W}_{up} \in \mathbb{R}^{d \times r_a},
    \end{gathered}
    \label{eq:adapter_bottleneck2}
\end{equation}

\noindent with bottleneck dimension $r_a \ll d$ and GELU activation.

\subsection{Federated Adaptation Protocol}
\label{apx:adapter:protocol}

This adaptation stage precedes the standard FedIAL continual learning process and is executed only once, before incremental learning begins. It consists of standard iterative federated optimization over the adapter parameters $\Phi = \{\mathbf{W}_{down}^{(\ell)}, \mathbf{W}_{up}^{(\ell)}\}_{\ell=1}^{L}$, with all ViT-B/16 weights frozen throughout. This is the only stage of the pipeline in which iterative gradient-based optimization is used; classifier learning across all subsequent tasks follows the closed-form RRLS formulation of \Cref{sec:method} without modification.

Each client $i$ minimizes cross-entropy loss over its local task-1 partition using a temporary linear classification head attached to the CLS token. Only $\Phi_i$ and the temporary head receive gradient updates; the backbone is not touched.

After local training, each client transmits $\Phi_i$ to the coordinator. Adapter training is restricted to task~1. Over its communication rounds, local adapter updates are aggregated with FedAvg \cite{mcmahan_communication-efficient_2017}:

\begin{equation}
    \Phi_g = \sum_{i=1}^{N} \frac{n_i}{n}\,\Phi_i,
    \label{eq:fedavg_adapter}
\end{equation}

\noindent where $n_i$ is the local sample count and $n = \sum_i n_i$. After the last round of task~1, the global adapter $\Phi_g$ is broadcast to all clients and frozen, and the temporary SGD head is discarded.

From task~2 onward, the composed extractor $F_{\theta,\Phi_g}$, the frozen ViT-B/16 with globally aligned adapters, produces all embeddings. This extractor is frozen for the remainder of training, and FedIAL's two-phase RRLS protocol (Algorithm~\ref{alg:aggregation}) applies without modification. No further gradient computation or gradient communication occurs.

\subsection{Implementation Details}
\label{apx:adapter:impl}

We use the ViT-B/16 checkpoint pretrained on ImageNet-21K and fine-tuned on ImageNet-1K, obtained from the \texttt{timm} library. Local training runs for 20 communication rounds and 2 epochs per client with batch size 128, SGD optimizer (momentum 0.9), initial learning rate $0.1$ with cosine annealing to $0.01$. Adapter dimension ($r_a$) is 64 and scale ($\gamma$) is set to 0.3 (based on \cite{chen2022adaptformer}).

\end{document}